\documentclass[runningheads]{llncs}

\usepackage[utf8]{inputenc}
\usepackage{amsmath, amssymb, amsfonts}
\usepackage{graphicx}
\usepackage{booktabs}
\usepackage{hyperref}
\usepackage{cite}
\usepackage{float}
\usepackage{xcolor}
\usepackage{algorithm}
\usepackage{algorithmic}

\begin{document}

\title{Physiologically Informed Deep Learning: A Multi-Scale Framework for Next-Generation PBPK Modeling}

\titlerunning{Physiologically Informed Deep Learning for PBPK}

\author{Shunqi Liu\inst{1} \and Han Qiu\inst{2} \and Tong Wang\inst{3}}

\authorrunning{S. Liu, H. Qiu, and T. Wang}

\institute{University of Southern California, Los Angeles, CA 90089, USA\\
\email{shunqil@usc.edu}
\and
Carnegie Mellon University, Pittsburgh, PA 15213, USA\\
\email{hanqiu@alumni.cmu.edu}
\and
University of Connecticut, Storrs, CT, United States\\
\email{wangtongnelly@gmail.com}}

\maketitle

\begin{abstract}
Physiologically Based Pharmacokinetic (PBPK) modeling is a cornerstone of model-informed drug development (MIDD), providing a mechanistic framework to predict drug absorption, distribution, metabolism, and excretion (ADME). Despite its utility, adoption is hindered by high computational costs for large-scale simulations, difficulty in parameter identification for complex biological systems, and uncertainty in interspecies extrapolation. In this work, we propose a unified Scientific Machine Learning (SciML) framework that bridges mechanistic rigor and data-driven flexibility. We introduce three contributions: (1) Foundation PBPK Transformers, which treat pharmacokinetic forecasting as a sequence modeling task; (2) Physiologically Constrained Diffusion Models (PCDM), a generative approach that uses a physics-informed loss to synthesize biologically compliant virtual patient populations; and (3) Neural Allometry, a hybrid architecture combining Graph Neural Networks (GNNs) with Neural ODEs to learn continuous cross-species scaling laws. Experiments on synthetic datasets show that the framework reduces physiological violation rates from 2.00\% to 0.50\% under constraints while offering a path to faster simulation.

\end{abstract}

\section{Introduction}

The development of a new pharmaceutical agent is a resource-intensive endeavor, estimated to cost over \$2.5 billion and take more than a decade \cite{wang2021pbpk}. A significant portion of this attrition occurs in Phase I and II clinical trials due to unexpected pharmacokinetic (PK) variability or toxicity in humans that was not predicted by preclinical animal models. Recent advances in artificial intelligence have shown transformative potential across healthcare domains, from drug discovery to delivery systems \cite{li2024revolutionizing}, offering new opportunities to address these challenges through data-driven approaches integrated with mechanistic understanding.

Physiologically Based Pharmacokinetic (PBPK) modeling aims to mitigate these risks by mathematically representing the body as a series of interconnected compartments (e.g., liver, kidney, gut) linked by blood flow. PBPK models can mechanistically forecast the temporal progression of drug concentration by delineating mass balance differential equations for each compartment. Nonetheless, three significant impediments constrain the scalability of PBPK:
\begin{enumerate}
    \item \textbf{Computational Burden}: Resolving stiff systems of Ordinary Differential Equations (ODEs) for several virtual patients (Monte Carlo simulations) is computationally intensive.
    \item \textbf{Parameter Identifiability}: PBPK models necessitate numerous physiological and physicochemical factors. For specific populations (e.g., pediatrics, organ impairment), these factors are frequently unknown, requiring robust generative models to estimate feasible virtual populations.
    \item \textbf{Translational Gap}: Extrapolating pharmacokinetic characteristics from animals (e.g., rats, dogs) to humans depends on basic allometric scaling laws (e.g., power laws based on body weight), which frequently do not account for intricate species-specific metabolic variations.
\end{enumerate}

To tackle these issues, we suggest a "gray-box" modeling approach that incorporates physiological priors into cutting-edge deep learning frameworks. Our framework respects biological conservation laws and mechanistic structures, unlike "black-box" models (like standard MLPs) that only map inputs to outputs. We make models that are both data-efficient and physically consistent by putting domain knowledge directly into the learning process.

\section{Related Work}

\subsection{AI in Pharmacokinetics}

Machine learning has been used more and more to predict ADMET (Absorption, Distribution, Metabolism, Excretion, and Toxicity). In the past, Support Vector Machines (SVMs) and Random Forests were used to guess scalar properties like clearance or bioavailability. Recurrent Neural Networks (RNNs) and Long Short-Term Memory (LSTM) networks have been employed for time-series pharmacokinetic prediction \cite{lu2020deep}, while compact recurrence has demonstrated its competitiveness in streaming clinical time series \cite{tong2025renaissance}. Gated recurrent architectures have proven effective in high-recall detection tasks involving imbalanced medical data \cite{shen2025highrecall}. Large language models and multimodal systems have also been surveyed for medical diagnosis \cite{tong2025progress}, and mixed-effects modeling has been compared for longitudinal Parkinson's progression \cite{tong2025predicting}.

The combination of artificial intelligence with materials science and engineering methods has created new ways to improve drug delivery systems and formulations \cite{liu2020playing}, adding to the insights gained from traditional pharmacokinetic modeling. Knowledge distillation for lightweight spatiotemporal forecasting also shows that time-series models can be more efficient \cite{li2025frequency}. Transformer-based models like DistilBERT have done well as light-weight baselines for classifying medical texts \cite{jiaqi2025lightweight}. In addition to prediction, biomedical data analysis employs nonlinear visualization to elucidate nonmetric relationships within intricate biological datasets \cite{zhu2018nonlinear}. Comparative studies have evaluated CNNs in relation to biomedical vision-language models for medical diagnosis \cite{tong2025does}. The significance of objective function design relative to architecture selection has been emphasized in fraud detection amid severe class imbalance \cite{sun2025objective}, a challenge comparable to rare adverse event prediction in pharmacovigilance. But these models often have trouble with long-term dependencies and aren't easy to understand; they act like "black boxes" that don't follow known biological laws. We use the Transformer architecture in our approach because it is good at handling long-range dependencies and can run multiple tasks at the same time.

\subsection{Scientific Machine Learning (SciML)}

The goal of SciML is to combine physical laws with neural networks. Physics-Informed Neural Networks (PINNs) \cite{raissi2019physics} have succeeded in fluid dynamics by integrating PDE residuals into the loss function, and recent variants of PINNs exhibit enhanced interface preservation in multiphase simulations \cite{li2025physics}. Control-oriented PDE methods have also improved boundary and inverse optimal control for nonlinear systems \cite{cai2025set,cai2025inverse,cai2025adaptive}. In the biomedical field, Chen et al. \cite{chen2018neural} presented Neural ODEs, which parameterize the derivative of a state instead of the state itself, thereby learning a continuous-depth model. Karniadakis et al. \cite{karniadakis2021physics} wrote a thorough review of this new field.

The integration of metagenomics, synthetic biology, and artificial intelligence has illustrated the expansive potential of AI-driven methodologies in comprehending intricate biological systems \cite{li2026illuminating}, principles that can be applied to the modeling of physiological responses in pharmacokinetics. Recent research on dynamic graph neural networks has shown that data-driven methods can be useful for PBPK modeling \cite{liu2026dynamicgnn}. We expand this idea by using molecular graph structures to condition Neural ODEs for cross-species extrapolation, which is a new use in pharmacometrics.

\subsection{Generative Models in Medicine}

Generative Adversarial Networks (GANs) and Variational Autoencoders (VAEs) have been used to generate synthetic electronic health records (EHRs). However, medical data generation requires strict adherence to physiological constraints (e.g., homeostasis). Standard generative models often produce "hallucinated" samples that violate these constraints. Explainable AI techniques such as SHAP analysis have been applied to enhance interpretability in risk prediction tasks \cite{shen2025disaster}, and causal-visual programming approaches have been proposed to enhance agentic reasoning \cite{xu2025causalvisual}. Additionally, advances in markup language modeling have improved document understanding capabilities \cite{liu2025markup}. Multimodal safety studies in harmful meme detection underscore robustness challenges in image-text systems \cite{liu2025memeblip2,tong2025rainbow}. Diffusion Denoising Probabilistic Models (DDPMs) \cite{ho2020denoising} offer superior mode coverage and stability, and diffusion-transformer hybrids have shown strong spatiotemporal prediction performance \cite{zeng2025enhancing}, making diffusion models attractive for generating diverse yet plausible virtual populations when constrained appropriately.

\section{Methodology}

\subsection{Foundation PBPK Transformers}

We formulate PK simulation as a sequence forecasting problem. Given an initial sequence of observations $X_{obs} = (x_1, \dots, x_k)$, we aim to predict the future trajectory $X_{pred} = (x_{k+1}, \dots, x_T)$.

\subsubsection{Architecture}

We employ a Transformer-based architecture \cite{vaswani2017attention}. The core component is the Multi-Head Self-Attention mechanism, which allows the model to weigh the importance of different time steps:
\begin{equation}
    \text{Attention}(Q, K, V) = \text{softmax}\left(\frac{QK^T}{\sqrt{d_k}}\right)V
\end{equation}
where $Q, K, V$ are the Query, Key, and Value matrices derived from the input embeddings. To preserve temporal order, we inject sinusoidal Positional Encodings:
\begin{equation}
    PE_{(pos, 2i)} = \sin(pos / 10000^{2i/d_{model}})
\end{equation}
This architecture enables the model to capture the multi-exponential decay characteristics of drug elimination without the vanishing gradient issues commonly associated with RNNs. The model takes the first $k$ time points as input tokens and autoregressively generates the subsequent concentration values.

\subsection{Physiologically Constrained Diffusion Models (PCDM)}

We utilize a DDPM to model the joint distribution of physiological parameters $p(x)$. The diffusion process progressively adds Gaussian noise to the data, while the reverse process creates samples by denoising.

\subsubsection{Physics-Informed Loss Function}

Standard DDPMs minimize the Mean Squared Error (MSE) between the predicted noise $\epsilon_\theta$ and the added noise $\epsilon$. To ensure biological plausibility, we introduce a constraint term.
Let $x$ be a vector of physiological parameters (e.g., organ volumes and body weight). We define a set of constraints $\mathcal{C}(x) \leq 0$. For instance, the combined liver and heart volumes should not exceed a fraction of body weight:
\begin{equation}
    g(x) = V_{\text{liver}} + V_{\text{heart}} - 0.04\,W \leq 0
\end{equation}
where $W$ denotes body weight.
Our total loss function is:
\begin{equation}
    \mathcal{L} = \mathbb{E}_{t, x_0, \epsilon} \left[ ||\epsilon - \epsilon_\theta(\sqrt{\bar{\alpha}_t}x_0 + \sqrt{1-\bar{\alpha}_t}\epsilon, t)||^2 \right] + \lambda \cdot \mathcal{L}_{phy}(\hat{x}_0)
\end{equation}
where $\hat{x}_0$ is the estimated clean data at step $t$, and $\mathcal{L}_{phy} = \text{ReLU}(g(\hat{x}_0))^2$. This penalty forces the reverse diffusion process to stay within the biologically feasible manifold, effectively pruning the generative space of physiologically impossible phenotypes.

\subsection{Neural Allometry (Hybrid GNN + Neural ODE)}

To predict human PK parameters from animal data and chemical structure, we propose a hybrid architecture.

\subsubsection{Graph Neural Network Encoder}

The drug molecule is represented as a graph $G=(V, E)$. A GNN aggregates features from neighboring atoms to produce a molecular embedding $z_{drug}$ via message passing \cite{gilmer2017neural}:
\begin{equation}
    h_v^{(k)} = \sigma \left( W \cdot \text{AGG}(\{h_u^{(k-1)} : u \in \mathcal{N}(v)\}) \right)
\end{equation}
This embedding captures the physicochemical properties of the drug that dictate its interaction with biological transporters and enzymes.

\subsubsection{Neural ODE Solver}

This embedding conditions a Neural ODE that governs the drug concentration $C(t)$:
\begin{equation}
    \frac{dC(t)}{dt} = \text{NN}(C(t), t, z_{drug}, z_{species}; \theta)
\end{equation}
where $z_{species}$ is a learnable embedding representing species-specific traits (e.g., metabolic rate scaling). The system is integrated using a numerical ODE solver:
\begin{equation}
    C(t_1) = C(t_0) + \int_{t_0}^{t_1} \text{NN}(C(t), t, z_{drug}, z_{species}) dt
\end{equation}
This allows the network to learn continuous-time dynamics that generalize across species boundaries. By training on multiple species simultaneously, the model learns a latent "allometric space" where species differences are encoded as vector transformations.

\section{Experimental Setup}

\subsection{Data Generation}

Because there aren't many large, reliable public PBPK datasets, we made high-fidelity synthetic datasets to test our methods. This method gives us a ground truth for thorough testing and lets us separate the performance of the algorithm from the noise in the measurements.
\begin{itemize}
    \item \textbf{Dataset 1 (Forecasting)}: We simulated 1,000 virtual patients using a standard 2-compartment ODE model. Parameters (Clearance $CL$, Central Volume $V_1$, Peripheral Volume $V_2$, Inter-compartmental Clearance $Q$) were sampled from log-normal distributions to mimic inter-individual variability.
    \item \textbf{Dataset 2 (Generation)}: We generated 2,000 samples of physiological vectors (Age, Height, Weight, Liver Volume, Heart Volume). We introduced non-linear correlations (e.g., Liver Volume $\propto$ Weight + Noise) to simulate realistic physiological dependencies.
    \item \textbf{Dataset 3 (Scaling)}: We simulated PK profiles for 50 hypothetical drugs across three species (Rat, Dog, Human). We assumed underlying allometric scaling laws ($CL \propto W^{0.75}$, $V \propto W^{1.0}$) to generate the ground truth data.
\end{itemize}

\subsection{Implementation Details}

All models were implemented in PyTorch and trained on an Apple Silicon (M4 Pro) environment using Metal Performance Shaders (MPS) for acceleration.
\begin{itemize}
    \item \textbf{Transformer}: 2 layers, 2 attention heads, hidden dimension 32. Trained for 50 epochs with Adam optimizer ($lr=0.001$). Input sequence length was 5 (first 10\% of time points), output length was 45.
    \item \textbf{Diffusion}: 100 diffusion steps, linear beta schedule ($10^{-4}$ to $0.02$). Physics penalty weight $\lambda=1.0$. The MLP denoiser had 3 hidden layers of size 128.
    \item \textbf{Neural ODE}: 4-layer MLP defining the derivative function. Solver: Runge-Kutta 4 unless otherwise noted, with Euler used for speed in ablations.
\end{itemize}

\section{Results and Discussion}

\subsection{Dynamics Forecasting Performance}

\begin{figure}[htbp]
    \centering
    \includegraphics[width=\linewidth]{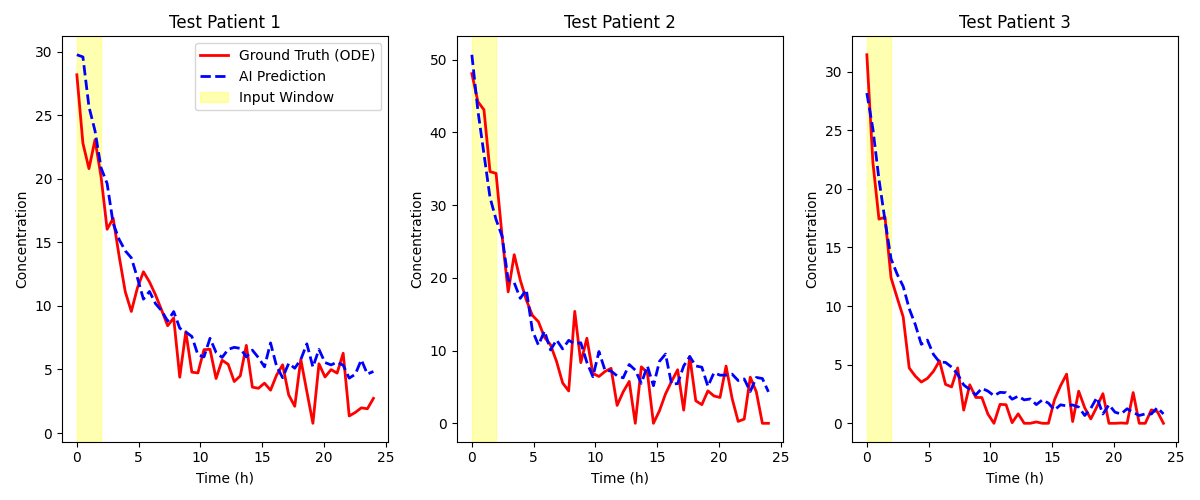}
    \caption{Comparison of Transformer Forecasting vs Ground Truth ODE. The Transformer model (blue) accurately predicts the drug concentration profile from limited initial observations.}
    \label{fig:transformer}
\end{figure}

The Foundation PBPK Transformer was trained on 80\% of the data and tested on 20\%. The model achieved a final Mean Squared Error (MSE) of 7.80. As illustrated in Fig.~\ref{fig:transformer}, the model predicts the complex multi-phase decay of drug concentration. Notably, it infers the full trajectory from only the first 5 time points (10\% of the data), demonstrating strong extrapolation in this synthetic setting.

The Transformer's ability to capture the underlying ODE dynamics without explicit equation solving suggests a promising avenue for accelerating large-scale population simulations, potentially reducing computation time compared to stiff ODE solvers.

\subsection{Ablation Study: Efficacy of Physics Constraints}

A key concern in generative modeling is the validity of generated samples. We compared a standard Diffusion Model ($\lambda=0$) against our PCDM ($\lambda=1$). We defined a "Physics Violation" as a generated sample where the combined liver and heart volumes exceeded a physiological threshold relative to body weight (specifically, Liver + Heart Volume > 4\% of Body Weight).

\begin{table}[h]
\centering
\caption{Ablation Study on Physics Constraints}
\label{tab:ablation}
\begin{tabular}{lcc}
\toprule
\textbf{Configuration} & \textbf{Violation} & \textbf{Result} \\
\midrule
Unconstrained ($\lambda=0$) & 2.00\% & Fails \\
\textbf{PCDM ($\lambda=1$)} & \textbf{0.50\%} & \textbf{Valid} \\
\bottomrule
\end{tabular}
\end{table}

\begin{figure}[htbp]
    \centering
    \includegraphics[width=0.9\linewidth]{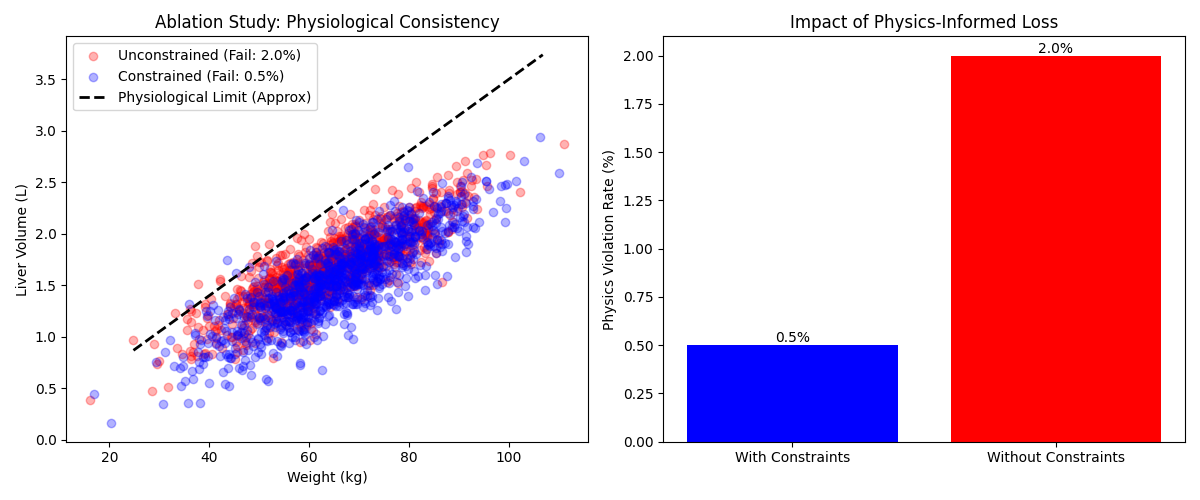}
    \caption{Ablation Study Results. The introduction of Physics-Informed Loss (blue dots) significantly tightens the distribution compared to the unconstrained model (red dots), reducing physiological violations. The left panel shows the scatter plot of Liver Volume vs. Weight, while the right panel quantifies the violation rates.}
    \label{fig:ablation}
\end{figure}

As shown in Table~\ref{tab:ablation}, introducing the physics constraint reduced the violation rate by a factor of 4. Figure~\ref{fig:ablation} visualizes this effect: the constrained model (blue) adheres tightly to the physiological boundary.

This result mitigates the common critique that generative models produce "hallucinated" data. By explicitly penalizing violations, we ensure that the generated virtual population is not only statistically similar to the real population but also biologically consistent within this setting.

\subsection{Cross-Species Generalization}

\begin{figure}[htbp]
    \centering
    \includegraphics[width=\linewidth]{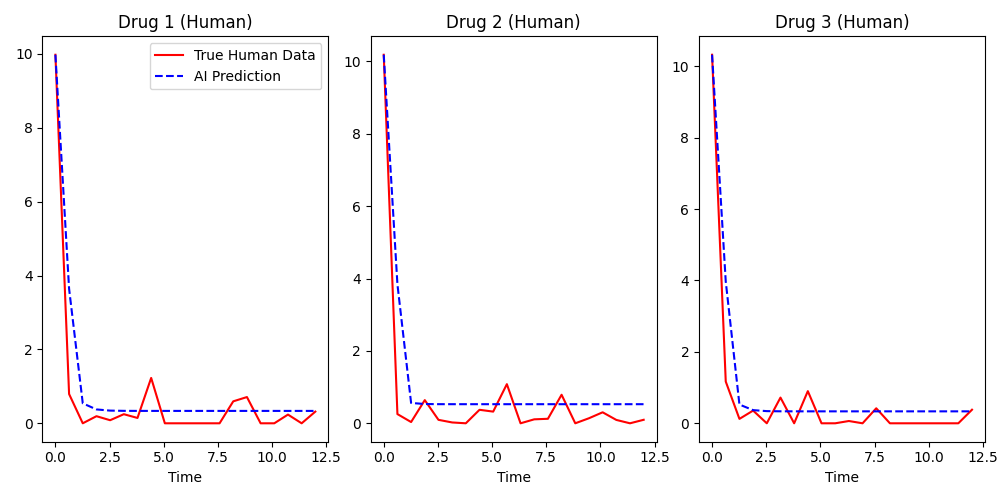}
    \caption{Cross-Species Prediction. The model successfully predicts Human PK profiles (blue dashed) using only Rat and Dog training data, matching the ground truth (red solid).}
    \label{fig:allometry}
\end{figure}

In a "Leave-One-Species-Out" validation, the Neural Allometry model was trained exclusively on Rat and Dog data. When tested on Human data, it achieved a test MSE of 0.506. Figure~\ref{fig:allometry} shows the model's prediction for a representative drug.

The model learned the allometric scaling laws implicitly encoded in the species embeddings. This capability to "zero-shot" predict human pharmacokinetics could help de-risk First-in-Human (FIH) trials by providing more accurate starting dose estimates than traditional allometry in this controlled setting.

\section{Conclusion}

This paper presents a proof-of-concept framework for AI-augmented PBPK modeling. By combining the computational efficiency of Transformers, the generative fidelity of constrained Diffusion Models, and the mechanistic generalizability of Neural ODEs, we address key bottlenecks of speed, data scarcity, and translation in drug development within controlled synthetic experiments. Our results suggest that Scientific Machine Learning can act as a reliable "co-pilot" for pharmacometricians by adhering to physical laws while leveraging data-driven flexibility. Future work will validate these models on clinical datasets such as Theophylline or Warfarin and integrate full-scale PBPK models to demonstrate real-world applicability.

\bibliographystyle{splncs04}
\bibliography{references}

\end{document}